\documentclass[journal,a4paper]{IEEEtran}
\IEEEoverridecommandlockouts
\usepackage{cite}
\usepackage{amsmath,amssymb,amsfonts}
\usepackage{algorithmic}
\usepackage{graphicx}
\usepackage{textcomp}
\usepackage{csquotes}
\usepackage{xcolor}
\usepackage{adjustbox}
\usepackage{hyperref}
\usepackage{multirow}
\def\BibTeX{{\rm B\kern-.05em{\sc i\kern-.025em b}\kern-.08em
    T\kern-.1667em\lower.7ex\hbox{E}\kern-.125emX}}
\DeclareMathOperator{\argmin}{argmin}
\newcommand{\R}{\ensuremath{\mathbb{R}}}
\begin{document}

\title{Set-Transformer BeamsNet for AUV Velocity Forecasting in Complete DVL Outage Scenarios}

\author{\IEEEauthorblockN{Nadav Cohen\IEEEauthorrefmark{1}, Zeev Yampolsky and Itzik Klein}\\
\IEEEauthorblockA{\textit{The Hatter Department of Marine Technologies,} \\
\textit{Charney School of Marine Sciences, University of Haifa}\\
Haifa, Israel}

\thanks{\IEEEauthorrefmark{1}Corresponding author: N. Cohen (email: ncohe140@campus.haifa.ac.il).}}

\maketitle

\begin{abstract}
Autonomous underwater vehicles (AUVs) are regularly used for deep ocean applications. Commonly, the autonomous navigation task is carried out by a fusion between two sensors: the inertial navigation system and the Doppler velocity log (DVL).  The DVL operates by transmitting four acoustic beams to the sea floor, and once reflected back, the AUV velocity vector can be estimated. However, in real-life scenarios, such as an uneven seabed, sea creatures blocking the DVL's view and,  roll/pitch maneuvers, the acoustic beams' reflection is resulting in a scenario known as DVL outage. Consequently, a velocity update is not available to bind the inertial solution drift.
To cope with such situations, in this paper, we leverage our BeamsNet framework and propose a Set-Transformer-based BeamsNet (ST-BeamsNet) that utilizes inertial data readings and previous DVL velocity measurements to regress the current AUV velocity in case of a complete DVL outage. The proposed approach was evaluated using data from  experiments held in the Mediterranean Sea with the Snapir AUV and was compared to a moving average (MA) estimator. Our ST-BeamsNet estimated the AUV velocity vector with an 8.547\% speed error, which is 26\% better than the MA approach.        

\end{abstract}


\section{Introduction}
Autonomous underwater vehicles (AUVs) are used in different applications.  Autonomous navigation with sufficient accuracy is one crucial aspect allowing the AUV to successfully complete its task. Therefore, the AUV is equipped with several navigation-enabling sensors \cite{jain2015review}. Often, the inertial navigation system (INS) and Doppler velocity log (DVL) are used to provide the navigation solution \cite{yonggang2013tightly,yao2020simple,klein2015observability}. 
%
The INS  consists of inertial sensors that measure the specific force and angular velocity vectors. These  are then integrated to provide the navigation solution, namely the AUV's  position, velocity, and orientation. \cite{titterton2004strapdown,groves2015principles}.However, due to the system's nature, the navigation solution accumulates error that needs to be corrected using an external aiding sensor \cite{farrell2007gnss,shin2002accuracy} such as a DVL.\\
The DVL transmits four acoustics beams to the seafloor that are reflected back to the sensor. Through this process, known as bottomlock, beam velocity measurements are obtained. It has been shown that DVL provides high-accuracy velocity estimates and is, therefore, considered a reliable sensor ~\cite{wang2019novel}. To further improve the DVL velocity estimate, we introduced BeamsNet, an end-to-end deep learning framework to estimate the AUV velocity vector using DVL data or inertial/DVL information \cite{cohen2022beamsnet}. However, in real-world scenarios, not all the DVL beams are reflected back due to several reasons, such as uneven seabed, sea creatures blocking the sensor's view, and during pitch/roll maneuvers \cite{eliav2018ins,klein2020continuous}. If fewer than three beam  measurements are available the DVL cannot provide the AUV velocity vector and  consequently the AUV would usually abort its mission and be forced back to the surface.\\
To obtain continuous DVL data flow, several solutions have been introduced addressing both partial (less than three beams) and complete (no beams) DVL outages.  In partial DVL scenarios, a virtual beam was generated using past measurements and additional data \cite{tal2017inertial}. Further development expanded this approach by considering the cross-covariance matrix of the correlated INS and DVL noise \cite{eliav2018ins}. Later, a non-linear filter was used for INS/DVL fusion using partial DVL measurements \cite{liu2018ins}. Recently, data-driven approaches were employed for navigation purposes \cite{klein2022} allowing for partial DVL measurements learning approach to emerge. Furthermore, a long short-term memory (LSTM) has been suggested as a model to compensate when one beam is missing \cite{yona2021compensating}, and later using least squares support vector machine-aided virtual beam construction algorithm \cite{yao2022Virtual}. Previously, we addressed a scenario of two missing DVL beams using a modified BeamsNet architecture \cite{cohen2022libeamsnet}. \\
To address complete DVL outages an analytical closed-form solution was developed to estimate  the AUV velocity, based on
past DVL measurements \cite{klein2020continuous,klein2022estimating}. Later, a nonlinear autoregressive model based on INS and DVL measurements was suggested and evaluated using a ship \cite{li2021underwater}.\\
This paper proposes, Set-Transformer based BeamsNet (ST-BeamsNet), a novel solution to handle complete DVL outages. ST-BeamsNet utilizes past AUV velocities and inertial data to forecast the current AUV velocity vector. In the first phase, the output of the two sensors goes through a patch-embedding layer that extracts features, and then utilizes the attention mechanism to learn data dependencies, disregarding their distance in the input. Later, the data is  aggregated using a learnable aggregation block and finally, the features and dependencies from both sensors are linked  to create the AUV velocity vector.  ST-BeamsNet was evaluated using data from  experiments in the Mediterranean Sea using the University of Haifa's Snapir AUV. Approximately four hours of recorded IMU and DVL data were collected and used to train, validate, and test our suggested approach.\\   
The rest of the paper is organized as follows: Section \ref{DVL} introduces the DVL equations and error models. Section \ref{LB} discusses the model's learning blocks. Section \ref{BeamsNet} describes the proposed approach, ST-BeamsNet, while our sea experiments and dataset are presented in Section \ref{dataset}. The results are provided in Section \ref{Res}  and the conclusions are discussed in Section \ref{con}.
\section{DVL Velocity Updates}\label{DVL}
\begin{figure}[h]
	\centering
		\includegraphics[width=\columnwidth]{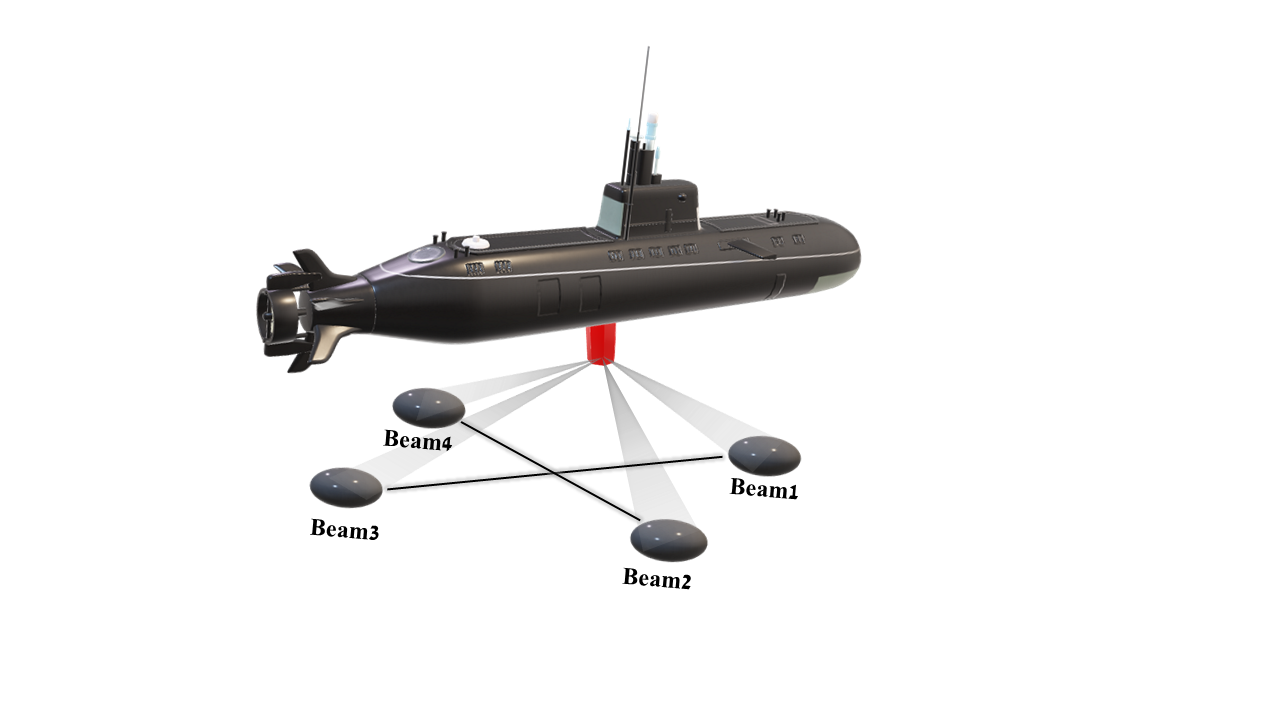}
	  \caption{A DVL transmits acoustic beams in \enquote*{$\times$} shape configuration".}\label{fig1}
\end{figure}

The DVL produces measurements based on the Doppler effect by transmitting four acoustic beams to the seafloor, commonly using  a Janus Doppler configuration, an\enquote*{$\times$} shape configuration, as can be seen in Fig. \ref{fig1}. By calculating the frequency shift between the transmitted beam and the one reflected to the sensor,  beam velocity measurements are obtained \cite{brokloff1994matrix}.\\
\\
In the \enquote*{$\times$} shape configuration the beams direction vectors are defined by
\begin{equation}\label{eqn:1}
    \centering
        \boldsymbol{beam}_{\dot{\imath}}=
        \begin{bmatrix} 
        \cos{\psi_{\dot{\imath}}}\sin{\theta}\quad
        \sin{\psi_{\dot{\imath}}}\sin{\theta}\quad
        \cos{\theta}
    \end{bmatrix}_{1\times3}
\end{equation} 
where the index $\dot{\imath}$ represents the beam number,  $\theta$ is the pitch angle, and the heading angle $\psi$ is 
\begin{equation}\label{eqn:2}
    \centering
        \psi_{\dot{\imath}}=(\dot{\imath}-1)\cdot\frac{\pi}{2}+\frac{\pi}{4}\;[rad]\;,\; \dot{\imath}=1,2,3,4
\end{equation}
The relationship between the AUV velocity in the body frame, $\boldsymbol{v}_{b}^{b}$, to the beam velocity measurements, $\boldsymbol{\upsilon}_{beam}$, can be expressed by a transformation matrix $\mathbf{H}$ \cite{liu2018ins}:  
\begin{equation}\label{eqn:3}
    \centering
        \boldsymbol{\upsilon}_{beam}=\mathbf{H}\boldsymbol{v}_{b}^{b} ,\quad
        \mathbf{H}=
        \begin{bmatrix} 
            \boldsymbol{beam}_{1}\\\boldsymbol{beam}_{2}\\\boldsymbol{beam}_{3}\\\boldsymbol{beam}_{4}\\
    \end{bmatrix}_{4\times3}
\end{equation} 
To solve \eqref{eqn:3}, a linear least squares (LS) filter is used:
\begin{equation}\label{eqn:5}
    \centering
        \hat{\boldsymbol{v}}_{b}^{b}=
        \underset{\boldsymbol{v}_{b}^{b}}{\argmin}{\mid\mid\boldsymbol{y}-\mathbf{H}\boldsymbol{v}_{b}^{b} \mid\mid}^{2}
\end{equation} 
where $\boldsymbol{y}$ is the measured beam velocity. 
The solution for the LS problem \eqref{eqn:5} is \cite{braginsky2020correction}:
\begin{equation}\label{eqn:6}
    \centering
        \hat{\boldsymbol{v}}_{b}^{b}=(\mathbf{H}^{T}\mathbf{H})^{-1}\mathbf{H}^{T}\boldsymbol{y}
\end{equation} 
The filter takes on two roles, filtering the errors and transforming the vector from the beam's coordinate frame to the AUV body frame.
\section{Learning Blocks}\label{LB}
We review here the basic learning blocks later used in our  proposed Set-Transformer BeamsNet approach. 
\subsection{Patch Embedding}
As suggested in \cite{dosovitskiy2020image}, the data is split into patches, flattened, and then linearly projected using a single one-dimensional convolution operation with a kernel size $\alpha$, stride size $\beta$, and  patch size $\gamma$. This patch embedding is represented by $\mathbf{x}_{p}\in \R^{N\times D}$,  where $N$ is the number of samples generated by the one-dimensional convolution layer and $D$ is the number of filters the layer uses and the latent space dimension fed to the Set-Transformers blocks.
\subsection{Attention}
The attention mechanism is defined by 
\begin{equation}
    \centering
        Attention(\mathbf{Q},\mathbf{K},\mathbf{V}) = Softmax(\frac{\mathbf{Q}\mathbf{K}^{T}}{\sqrt{d_{q}}})\mathbf{V} \in  \R^{n\times d_{v}}\label{eqn:7}
\end{equation}
where $n$ query vectors of size $d_{q}$ are represented by $\mathbf{Q}\in \R^{n\times d_{q}}$ and $n_{v}$ key - values pairs are $\mathbf{K}\in \R^{n_{v}\times d_{q}}$ and $\mathbf{V}\in \R^{n_{v}\times d_{v}}$, respectively,  $d_{v}$ is the corresponding values vector dimensions.  The Softmax function is \cite{goodfellow2016deep}
\begin{equation}
    \centering
        Softmax(\boldsymbol{Z})_{\dot\imath} = \frac{e^{Z_{\dot\imath}}}{\sum_{n=1}^{M} e^{Z_{\dot\jmath}}}\label{eqn:8}
\end{equation}
and $M$ is the number of vectors. 
\subsection{Multihead Attention}
Multiheads are defined to extend the attention mechanism. The multihead attention model projects query $Q$, keys $K$, and values $V$ vectors onto $h$ different heads and $\acute{d_{q}}$, $\acute{d_{q}}$, $\acute{d_{v }}$ dimensional vectors, respectively. The previously introduced mechanism \eqref{eqn:7} is applied to each $h$ projections.\\
Using multihead attention, the model can utilize information from different representation
subspaces and positions.\\
The model consists of an attention mechanism with trainable parameters $\mathbf{W}_{\dot\jmath}^{Q}$, $\mathbf{W}_{\dot\jmath}^{K}\in \R^{d_{q}\times \acute{d_{q}}}$ and $\mathbf{W}_{\dot\jmath}^{V}\in \R^{d_{v}\times \acute{d_{v}}}$
\begin{equation}
    \centering
        head_{\dot\jmath} =  Attention(\mathbf{Q}\mathbf{W}_{\dot\jmath}^{Q},\mathbf{K}\mathbf{W}_{\dot\jmath}^{K},\mathbf{V}\mathbf{W}_{\dot\jmath}^{V}),\; \dot\jmath = 1,...,h\label{eqn:9}
\end{equation}
The multiple heads are linked together in a series and multiplied with a trainable parameter $\mathbf{W}^{O}\in \R^{h\acute{d_{v }}\times d}$ resulting in the following: 
\begin{equation}
    \centering
        Multihead(\mathbf{Q},\mathbf{K},\mathbf{V})=concatenate(head_{1},..., head_{h})\mathbf{W}^{O}
\end{equation}
A common dimension hyper-parameters choice is $\acute{d_{q}}= \frac{d_{q}}{h}$, $\acute{d_{v}}= \frac{d_{v}}{h}$ and $d=d_{q}$.
\subsection{Set Attention Block}
The multihead attention block (MAB) is defined by modifying the encoder block of the Transformer introduced in \cite{vaswani2017attention} by removing the positional encoding and dropout operation. Given two matrices $\mathbf{X},\mathbf{Y} \in \R^{n\times d}$ the parameter $\mathbf{L}$ is defined as:
\begin{equation}
    \centering
      \mathbf{L} = LayerNorm(\mathbf{X}+ Multihead(\mathbf{X},\mathbf{Y},\mathbf{Y})) \label{eqn:11} 
\end{equation}
and the MAB is defined as
\begin{equation}
    \centering
      MAB(\mathbf{X},\mathbf{Y}) = LayerNorm(\mathbf{L}+ FFN(\mathbf{L})) \label{eqn:12} 
\end{equation}
were $FFN$ is a fully connected feed-forward network  built from a layer that expands the dimension to the size of $FFE$, the feed-forward expansion hyper-parameter. After applying a ReLU activation function, the second layer transforms it back to the original dimension
\begin{equation}
    \centering
     FFN(\mathbf{X}) = max(0,xW_{1}+b_{1})W_{2}+b_{s} \label{eqn:13} 
\end{equation}
The set attention block (SAB) performs self-attention between the elements in the set and can be defined using the MAB as follows: 
\begin{equation}
    \centering
     SAB(\mathbf{X}) := MAB(\mathbf{X},\mathbf{X}) \label{eqn:14} 
\end{equation}
\subsection{Pooling by Multihead Attention}
This block introduces a learnable aggregation scheme by defining a set of $k$ vectors of size $n$ and stacking them in a matrix $\mathbf{S}\in\R^{k\times n}$. To that end, by assuming that an encoder produces a set of features $\mathbf{Z}\in\R^{n\times d}$, then the pooling by multihead attention block (PMA) is defined as follows:
\begin{equation}
    \centering
     PMA_{k}(\mathbf{Z}) = MAB(\mathbf{S},FFN(\mathbf{Z})) \label{eqn:15} 
\end{equation}
where $k$ is a hyper-parameter.
\subsection{Overall Architecture}
The architecture is built from an encoder followed by a decoder. Given a matrix $\mathbf{X} \in \R^{n\times d}$ the encoder outputs the extracted features matrix  $\mathbf{Z}\in\R^{n\times d}$. The encoder is assembled by stacking SAB blocks \eqref{eqn:14}:
\begin{equation}
    \centering
     \mathbf{Z}=Encoder(\mathbf{X}) = SAB_{1}\circ...\circ SAB_{b}(\mathbf{X})\label{eqn:16}
\end{equation}\\
The decoder used the learnable aggregation to transform $\mathbf{Z}$ into a set of vectors that passes through a feed-forward network to receive the outputs \cite{lee2019set}.
\begin{equation}
    \centering
     Decoder(\mathbf{Z}) = \circ...\circ FFN_{b}(SAB(PMA_{k}(\mathbf{Z})))\in \R^{k\times d}\label{eqn:17}
\end{equation}
where $b$ is the number of stacked blocks.\\
To conclude, the hyper-parameters that need to be determined in the blocks above are:
\begin{itemize}
    \item $\alpha$ - Patch-Embedding kernel size
    \item $\beta$ - Patch-Embedding stride size
    \item $\gamma$ - Patch-Embedding patch size
    \item $D$ - Number of filters in the 1D convolution layer
    \item $d$ - Latent space dimension equaling to $D$
    \item $h$ - Number of attention heads
    \item $FFE$ - Feed forward expansion
    \item $b$ - Number of stacked SAB blocks
    \item $k$ - Number of trainable vectors for aggregation
\end{itemize}
\section{Set Transformer based BeamsNet}\label{BeamsNet}
\begin{figure*}[h]
\centering
\includegraphics[width=0.8\textwidth]{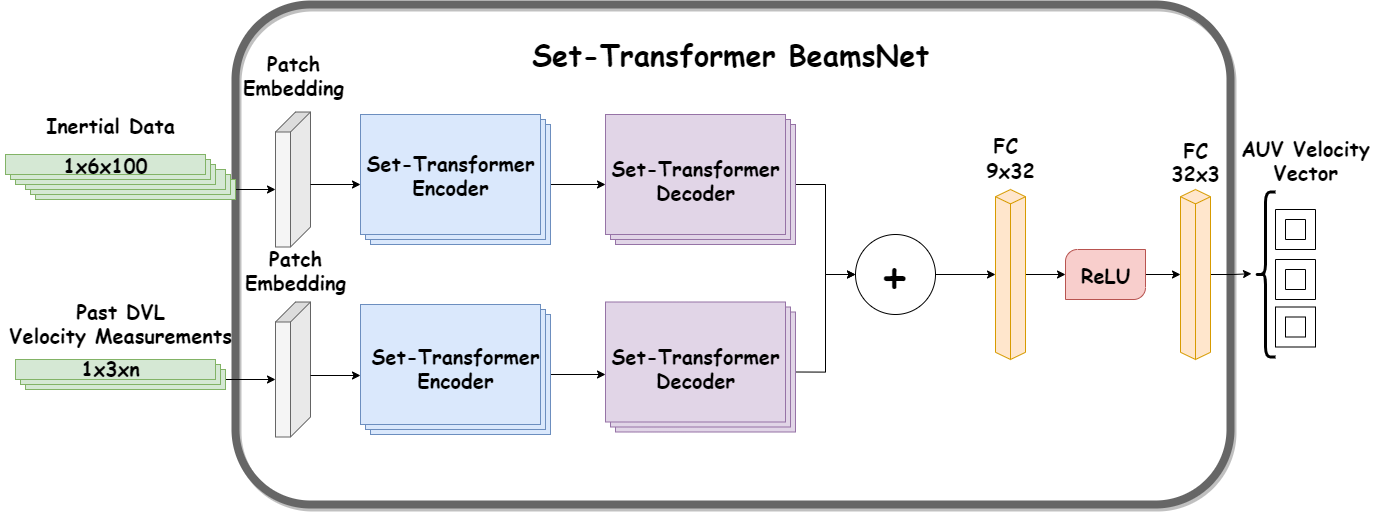}
\caption{A block diagram of the set-transformer based BeamsNet}\label{fig2}
\end{figure*}

Our proposed approach utilizes data from the inertial sensors, namely the accelerometers and gyroscopes, and $n$ past DVL velocity vector measurements.
The motivation behind the specific architecture is to learn features from each sensors' data using the patch embedding layer, which is later on inspected for dependencies using the multihead attention mechanism. Then, a learnable pooling aggregation is applied to obtain the most valuable features and dependencies, which are used to find links between the sensors using fully connected layers to regress the AUV velocity vector.
\\
The flow of information is as follows: first, the data goes through a patch-embedding layer, then it passes through $b$ blocks of encoders of decoders. Finally, the decoders' output from both sensors' data is combined and processed using two fully connected layers separated by the ReLU activation function. Since the network's purpose is to forecast the $n+1$ velocity vector, the loss function is the mean squared error (MSE). A block diagram showing our proposed approach is presented in Fig. \ref{fig2}.
 The hyper-parameters that were used in the set-transformer-based BeamsNet are presented in Table. \ref{table:1}.
\begin{table}[h]
\caption{ST-BeamsNet's hyper-parameters and values}
\centering
\begin{adjustbox}{width=\columnwidth}
\label{table:1}
\begin{tabular}{|c|c|c|c|c|c|c|c|c|}
\hline
Hyper-parameter & $\alpha$ & $\beta$ & $\gamma$ &  $D$ &\ $h$  & $ffe$ & $b$& $k$\\ \hline
Value           & 2 & 1 & 1 & 128 & 8 & 256 &2 & 3 \\ \hline
\end{tabular}
\end{adjustbox}
\end{table}
\section{Dataset} \label{dataset}
\begin{figure}[h]
	\centering
		\includegraphics[width=\columnwidth]{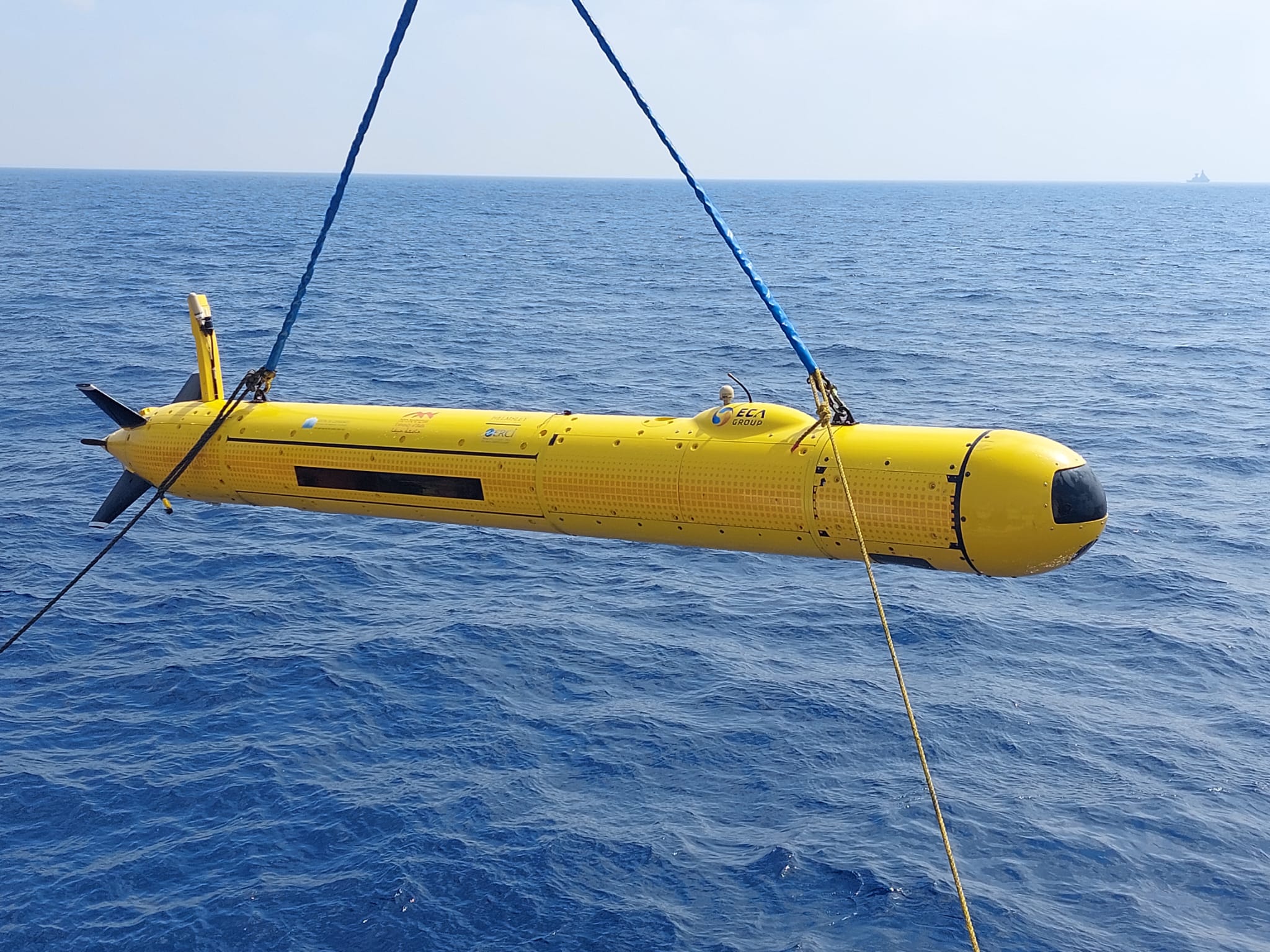}
	  \caption{The Snapir AUV before starting the experiment in the Mediterranean Sea. }\label{fig:snapir}
\end{figure}

The dataset was collected using the "Snapir" AUV,  an A18-D ECA GROUP mid-size AUV shown in Figure~\ref{fig:snapir}. Snapir's length and diameter are 5.5 [m] and  0.5 [m], respectively. Snapir can perform missions of up to 24 hours and reach 3000 [m] depth. The experiments took place in the Mediterranean Sea, and  data were collected from the AUV's inertial sensors (iXblue Phins Subsea IMU \cite{iXblue}0 and DVL (Teledyne RD Instruments, Navigator DVL \cite{Teledyne}) over nine  missions. The DVL has a sampling  rate of 1 [Hz], and  the inertial sensors sample at 100 [Hz]. The total duration of the recorded data is approximately four hours and include 13,886 DVL and 1,388,600 IMU measurements. The suggested network was trained on eight randomly chosen missions and tested on the remaining mission. \\
As the Snapir AUV carries only one DVL sensor, to mimic a scenario of a sensor that provides ground truth and a unit under test we added a noise term to the DVL measurements. To that end,  a scale factor of 0.7\%, a bias of 0.0001[m/s], and zero-mean white Gaussian noise of 0.042[m/s] were added. Once the data was created, a scenario of a complete DVL outage was created by removing the AUV velocity vector every four seconds.
\section{Results}\label{Res}
To evaluate the suggested approach, four metrics were used:
\begin{enumerate}
    \item  Root Mean Squared Error (RMSE)
    \begin{equation}\label{eqn:18}
    \centering
        RMSE(\boldsymbol{x}_{\dot\imath},\hat{\boldsymbol{x}}_{\dot\imath})=\sqrt{\frac{\sum_{\dot\imath=1}^{N}(\boldsymbol{x}_{\dot\imath}-\hat{\boldsymbol{x}}_{\dot\imath})^{2}}{N}}
\end{equation}
    \item  Mean Absolute Error (MAE)
    \begin{equation}\label{eqn:19}
    \centering
        MAE(\boldsymbol{x}_{\dot\imath},\hat{\boldsymbol{x}}_{\dot\imath})=\frac{\sum_{\dot\imath=1}^{N}|\boldsymbol{x}_{\dot\imath}-\hat{\boldsymbol{x}}_{\dot\imath}|}{N}
\end{equation}
    \item  The coefficient of determination ($R^{2}$)
    
\begin{equation}\label{eqn:20}
    \centering
        R^{2}(\boldsymbol{x}_{\dot\imath},\hat{\boldsymbol{x}}_{\dot\imath})=1- \frac{\sum_{\dot\imath=1}^{N}(\boldsymbol{x}_{\dot\imath}-\hat{\boldsymbol{x}}_{\dot\imath})^{2}}{\sum_{\dot\imath=1}^{N}(\boldsymbol{x}_{\dot\imath}-\bar{\boldsymbol{x}}_{\dot\imath})^{2}}
\end{equation}
    \item  The variance account for (VAF)
    
\begin{equation}\label{eqn:21}
    \centering
        VAF(\boldsymbol{x}_{\dot\imath},\hat{\boldsymbol{x}}_{\dot\imath})=[1-\frac{var(\boldsymbol{x}_{\dot\imath}-\hat{\boldsymbol{x}}_{\dot\imath})}{var(\boldsymbol{x}_{\dot\imath})}]\times100
\end{equation}
\end{enumerate}
The RMSE and MAE represnt the velocity error in units of [m/s], while R2 and VAF are unitless. An optimal solution would be if the VAF is $100$, the $R^{2}$ is $1$, and the RMSE and MAE are $0$ \cite{armaghani2021comparative}.
In the above equations: 
\begin{itemize}
    \item  $N$ is the number of samples
    \item  $\boldsymbol{x}_{\dot\imath}$ is the ground truth velocity vector norm of the DVL
    \item   $\hat{\boldsymbol{x}}_{\dot\imath}$ is the predicted velocity vector norm of the AUV given by the network
    \item  $\bar{\boldsymbol{x}}_{\dot\imath}$ is the mean of the ground truth velocity vector norm of the DVL
    \item $var$ stands for variance.
\end{itemize}
In addition, for a fair comparison, the Moving Average (MA) was calculated to predict the AUV velocity measurements based on $n=3$ past AUV velocity measurements \cite{smith2013digital}. Furthermore, the magnitude of the speed error was calculated by comparing the results to the mean of the ground truth velocity vector norm. The results are presented in Table \ref{table:2}.
\begin{table}[h!]
\caption{Comparison between ST-BeamsNet to the moving average method.}
\centering
   \begin{adjustbox}{width=\columnwidth}
\begin{tabular}{|c|c|c|}
\hline
Evaluation Metrics & \multicolumn{1}{l|}{ST-BeamsNet} & \multicolumn{1}{l|}{Moving-Average} \\ \hline
RMSE $[m/s]$       & 0.098                             & 0.134                       \\ \hline
RMSE $[\%]$        & 8.574                                 & 11.714                           \\ \hline
MAE $[m/s]$        & 0.064                             & 0.096                       \\ \hline
MAE $[\%]$         & 5.649                                & 8.395                          \\ \hline
$R^{2}$            & 0.978                             & 0.968                       \\ \hline
VAF                & 97.997                            & 96.971                      \\ \hline
\end{tabular}
\end{adjustbox}
    \label{table:2}
\end{table}\\
The mean of the ground truth velocity vector norm for the test set is $1.14$ [m/s].  These results indicate that the suggested approach can better forecast the AUV velocity vector, using  $n=3$ previous DVL velocity vector measurements and the inertial data readings, compared to the MA algorithm in each of the selected metrics. Besides obtaining a speed error of $8.571$\% and $5.649$\% with respect to the RMSE and MAE metrics, respectively, compared to the mean of the ground truth velocity vector norm, the set-transformer based BeamsNet outperforms the MA estimator and provides a $26\%$ improvement.
\begin{figure}[h]
	\centering
		\includegraphics[width=\columnwidth]{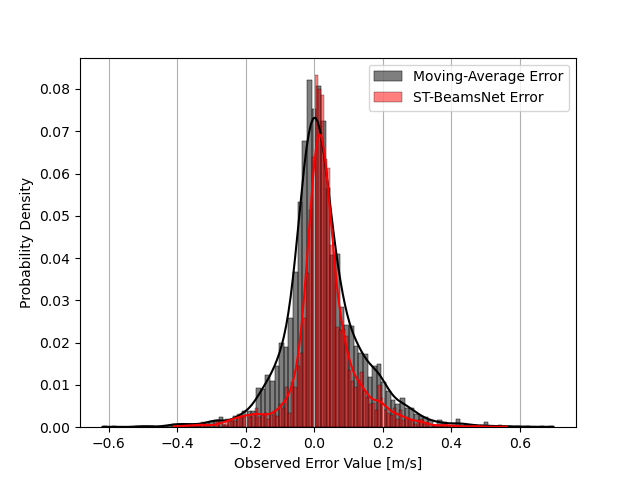}
	  \caption{A probability density error graph for the suggested approach (in red) and the MA method (in black). }\label{fig4}
\end{figure}\\
Fig. \ref{fig4} presents the probability density error graph for both methods. The ST-BeamsNet's mean error value is $0.022$[m/s], and the MA estimator mean error value is $0.028$[m/s]. Also, a difference in the error variance value was obtained where for the ST-BeamsNet approach the value is $0.008$[m/s] while for the MA method  $0.013$[m/s].   
\section{Conclusions}\label{con}
This paper introduced ST-BeamsNet, an end-to-end set-transformer based architecture to forecast to AUV velocity vector in a case of a complete DVL outage. The suggested approach utilizes data from the inertial sensors and past AUV velocity measurements to obtain the current velocity and, therefore, could prevent the navigation solution from drifting. By gathering data from an experiment using Snapir AUV in the Mediterranean Sea, the model was trained, tested, and  compared to an MA estimator.

The results show that ST-BeamsNet can forecast the current AUV velocity in the case of a complete DVL outage, yielding a speed error of 5.649\% and 8.54\% with respect to the MAE and RMSE metrics compared to the average speed, respectively. Furthermore, compared to the MA method, the suggested approach provides an improvement of 26\% and is characterized by a smaller variance error.  


\section*{Acknowledgments}
N.C. is supported by the Maurice Hatter Foundation and University of Haifa presidential scholarship for students on a direct Ph.D. track. 

\bibliographystyle{IEEEtran}
\bibliography{refs}

\end{document}